\documentclass[10pt,twocolumn,letterpaper]{article}

\usepackage{cvpr}
\usepackage{times}
\usepackage{epsfig}
\usepackage{graphicx}
\usepackage{amsmath}
\usepackage{amssymb}

\usepackage{bm}
\usepackage{subfig}
\usepackage{caption}
\usepackage{float}
\usepackage{multirow}
\usepackage{tabularx}
\usepackage{booktabs}
\usepackage{array}
\usepackage{tikz}
\usepackage{tikz-cd}
\usepackage{paralist}
\usepackage{xfrac}
\usepackage{mathtools,etoolbox}
\DeclarePairedDelimiterX{\abs}[1]{\lvert}{\rvert}{\ifblank{#1}{{}\cdot{}}{#1}}
\DeclarePairedDelimiterX{\inp}[2]{\langle}{\rangle}{#1, #2}

\usepackage[pagebackref=true,breaklinks=true,letterpaper=true,colorlinks,bookmarks=false]{hyperref}

\cvprfinalcopy 

\def\cvprPaperID{6363} 

\ifcvprfinal\fi
\begin{document}

\title{A-TVSNet: Aggregated Two-View Stereo Network for \protect\\
Multi-View Stereo Depth Estimation}

\makeatletter
\newcommand{\printfnsymbol}[1]{%
  \textsuperscript{\@fnsymbol{#1}}%
}
\makeatother

\author{Sizhang Dai\thanks{Equal contribution}\qquad Weibing Huang\printfnsymbol{1}\\
Shenzhen Evomotion Co., Ltd, China\\
{\tt\small szdai@evomotion.com\qquad whuang@evomotion.com}
}

\maketitle

\begin{abstract}

   We propose a learning-based network for depth map estimation from multi-view stereo (MVS) images. 
   Our proposed network consists of three sub-networks: 
   \begin{inparaenum}[1)]
      \item a base network for initial depth map estimation from an unstructured stereo image pair, 
      \item a novel refinement network that leverages both photometric and geometric information, and
      \item an attentional multi-view aggregation framework that enables efficient information exchange and integration among different stereo image pairs.
   \end{inparaenum}
   The proposed network, called A-TVSNet, is evaluated on various MVS datasets and 
   shows the ability to produce high quality depth map that outperforms competing approaches.
   Our code is available at \url{https://github.com/daiszh/A-TVSNet}. 
\end{abstract}


   \vspace{-0.2cm}
   \section{Introduction} \label{sec:intro}
   3-D reconstruction is a crucial problem in many fields of computer vision and computer graphics, \eg, augmented reality, CAD, medical imaging.  
   Multi-view stereo (MVS) is a commonly used approach in 3-D reconstruction. 
   Given a set of unstructured images and corresponding camera parameters, 
   MVS methods leverage underlying geometry information and reconstruct dense 3-D representation of a scene from all input views. 
   Although many efforts \cite{furukawa2010pmvs,galliani2015massively,goesele2007community,schonberger2016colmap} 
   have been devoted to improving the reconstruction quality, 
   state-of-the-art methods still suffer from artifacts and incompleteness caused by 
   low-textured regions, occlusions, non-Lambertian reflectance \etc in real-world scenes. 
   
   Recently, many studies \cite{huang2018deepmvs,im2019dpsnet,yao2018mvsnet} have applied convolution neural networks (CNNs) to the MVS task and shown promising results. 
   Most of these works can be seen as extensions of the CNN that handles the stereo matching problem \cite{chang2018psmnet,kendall2017gcnet,liang2018iresnet,mayer2016dispnet} 
   which aims to estimate the disparity map from a rectified image pair. 
   A typical two-view stereo algorithm performs (subsets of) the following four steps \cite{scharstein2002taxonomy}: 
   \begin{inparaenum}[1)]
      \item matching cost computation, 
      \item cost regularization, 
      \item disparity computation/optimization, 
      \item disparity refinement.
   \end{inparaenum}
   In CNN-based stereo matching algorithms, the matching cost computation is often implemented as 
   either a 3-D correlation volume between the two extracted feature maps across various disparity values \cite{liang2018iresnet,mayer2016dispnet}, 
   or a 4-D cost volume by concatenating feature maps of the left image and those of the horizontally displaced right image \cite{chang2018psmnet,kendall2017gcnet}. 
   While stereo matching considers only rectified image pairs, MVS needs to deal with the problem of varying camera poses.  
   To this end, the plane-sweep technique \cite{collins1996sweep,gallup2007sweep} is introduced in \cite{huang2018deepmvs,im2019dpsnet,yao2018mvsnet}. 
   In these works, the plane-sweep algorithm is used to warp the extracted feature maps of the neighboring images onto a series of successive virtual depth planes. 
   The multi-view matching cost is then conducted via a variance-based approach \cite{yao2018mvsnet, yao2019rmvsnet} or a concatenation-based approach \cite{huang2018deepmvs, im2019dpsnet}. 
   For the latter three steps (cost regularization, disparity computation and disparity refinement), 
   current CNN-based MVS methods are quite similar to those of the stereo matching: CNNs such as the U-Net \cite{ronneberger2015u} like architecture are used 
   in regularizing the cost volume and inferring the depth map, with optional refinement network \cite{yao2018mvsnet} to further improve the accuracy of the depth estimations. 
   
   Although many efforts have been made on designing a better MVS network in recent years,  
   current CNN-based methods do not seem to significantly outperform the traditional ones like \cite{galliani2015massively,schonberger2016colmap}. 
   We argue that it is because two important pieces of information, \ie, geometric consistency and multi-view information aggregation, 
   are either missing or insufficiently exploited in current works. 
   Geometric consistency is widely used in many traditional MVS algorithms \cite{schonberger2016colmap, xu2019multigeo} 
   to filter matching outliers and resolve ambiguities. 
   In the case of stereo matching, the geometric consistency degenerates to the left-right consistency 
   as is used in \cite{chabra2019stereodrnet} to refine the initial depth estimation. 
   While the left-right consistency is often merely considered as a post-processing step, 
   the geometric consistency plays a vital role in many conventional MVS algorithms: 
   COLMAP \cite{schonberger2016colmap} integrates it as a second step optimization given the initial depth estimation which is solely based on photometric clues, 
   Xu and Tao \cite{xu2019multigeo} present a multi-scale patch matching with geometric consistency guidance to constrain the depth optimization at finer scales. 
   If these steps were removed, a dramatic performance drop would be observed. 
   Surprisingly, the geometric consistency has not been exploited in any learning-based MVS method to the best of our knowledge. 
   
   Multi-view information aggregation is another key component that has not been efficiently exploited in learning-based methods. 
   \cite{choy20163d,kar2017lsm} treat MVS as a sequential problem and use recurrent neural networks (RNNs) to fuse the multi-view information. 
   Even though these works have shown promising results, recurrent architectures are order-sensitive 
   and unable to reconstruct consistent 3-D scene from different permutations of the same input image sequence \cite{vinyals2015order}.  
   Other learning-based methods tackle this problem using explicit aggregation operations. 
   Yao \etal \cite{yao2018mvsnet} propose a variance based cost to capture the second moment information for multi-view cost aggregation. 
   Variance of image features is a good indicator that facilitates the training process, 
   however, a lot of valuable feature information is discarded during the process of calculating variance. 
   \cite{huang2018deepmvs, im2019dpsnet} 
   use a max/mean pooling layer after the cost regularization to fuse information from different pairs. 
   Although the multi-view information is not prematurely discarded in these approaches, 
   simple pooling layer cannot capture the contextual information of different views 
   and bad estimations in one branch will often spoil the final result. 
   Moreover, unlike conventional methods, in which the multi-view information is constantly exchanged during the optimization process, 
   the multi-view aggregation in recent learning-based methods happens at one specific point 
   and thus is not designed for efficient information exchange among different views. 

   To overcome the above issues, we propose an aggregated two-view stereo network (A-TVSNet) for MVS depth estimation.  
   A-TVSNet regards the MVS problem as two subproblems: 
   \begin{inparaenum}[1)]
      \item to estimate depth from an unstructured two-view image pair, 
      \item to efficiently exchange and aggregate the information among multiple two-view instances. 
   \end{inparaenum}
   This decomposition is inspired by two essential differences between MVS and stereo matching: 
   \begin{inparaenum}[1)]
      \item the varying camera poses and 
      \item the varying number of images. 
   \end{inparaenum}
   Following this idea, we first build a two-view stereo network that takes an unstructured image pair as input and outputs the reference depth map. 
   To take advantages of the geometric information, the initial depth maps of both the reference image and its neighbor are estimated via the shared two-view network 
   to construct a geometric cost volume, which is then incorporated into a refinement module to obtain the refined result. 
   Secondly, we design a novel aggregation framework that enables efficient information summarization and exchange among multiple two-view networks. 
   A-TVSNet processes $N$ two-view networks in parallel and associates them using aggregation modules. 
   Unlike existing MVS methods, our framework allows the existence of multiple aggregation operations at flexible locations, 
   which enables repeated back-and-forth information exchanges among networks. 
   The $N$ individual local information flows are fused into a global one by a aggregation module, and then passed back to each individual networks. 
   Each of these $N$ two-view networks uses this shared knowledge together with its local knowledge for further computations. 
   
   
   To summarize our contributions in this work: 
   \begin{itemize}
      \setlength{\itemsep}{1mm}
      \setlength{\parsep}{0pt}
      \setlength{\parskip}{0pt}
	   \item
      We reformulate the MVS problem into two subproblems: 
      1) the unstructured two-view stereo matching, and 
      2) multi-view information aggregation. 
      Based on this decomposition, we propose A-TVSNet for MVS depth estimation.
      \item 
      In A-TVSNet, we adopt an end-to-end two-view stereo network that 
      leverages both photometric consistency and geometric consistency information. 
      \item 
      We design an order-invariant aggregation module that generalizes the work from \cite{Yang2018AttentionalAO} 
      to allow for stronger interaction among different information sources, 
	   and show how to apply this module to information aggregation in A-TVSNet. 
      \item
      Through extensive evaluations, we demonstrate the efficacy of our A-TVSNet for MVS depth estimation, 
      and our network achieves better performance than recent state-of-the-art learning-based methods. 
   \end{itemize}

\begin{figure*}[th]
   \begin{center}
   \includegraphics[width=\linewidth]
                     {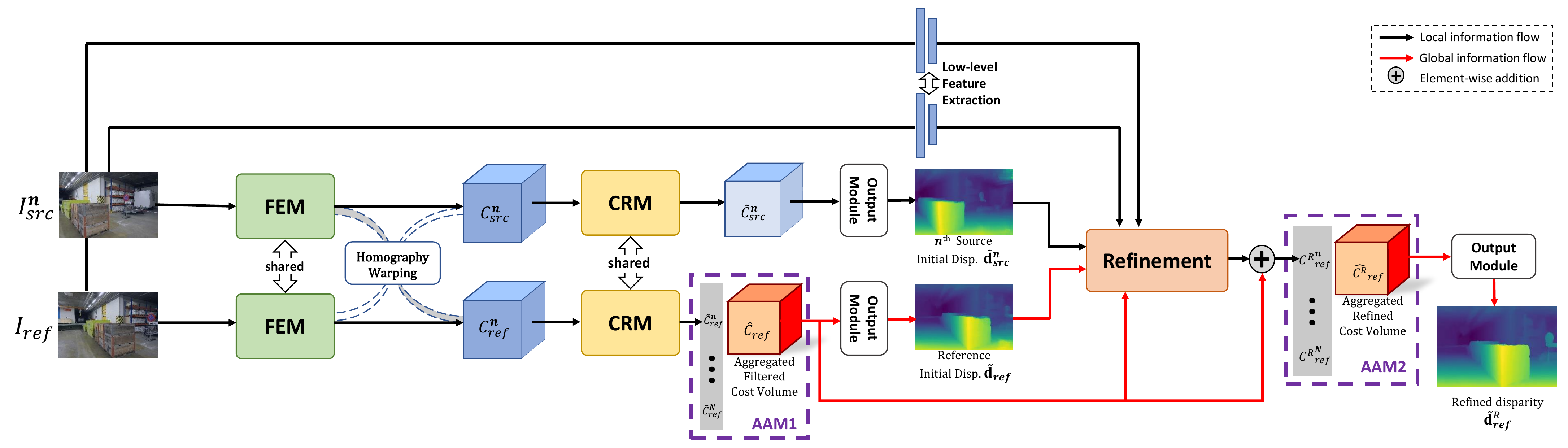}
   \end{center}
   \vspace{-0.2cm}
   \caption{Overview architecture of A-TVSNet. 
   Given a reference image and $N$ source images, each pair of images 
   $\left( I_{ref}, I^n_{src} \right)_{n=1}^{N}$ is processed in parallel by a shared two-view stereo network. 
   Two aggregation modules (AAM1 and AAM2) enable information exchanged and integration among these $N$ networks, 
   so that one unified estimate of the reference disparity map can be obtained as the output of our system. 
   } 
\label{fig:overview}
\vspace{-0.2cm}
\end{figure*}

\section{Related works} 

Estimation of depth from images has a long history in the literature, we refer readers to 
the exhaustive surveys of stereo matching \cite{scharstein2002taxonomy} and multi-view stereo \cite{furukawa2015tutorial}. 
As stated in \cite{yao2018mvsnet}, MVS methods can be divided into three categories: 
\begin{inparaenum}[1)]
   \item Point cloud based method \cite{furukawa2010pmvs,jancosek2011multi,lhuillier2005quasi,locher2016progressive}. 
   \item Volumetric based method \cite{ji2017surfacenet,kar2017lsm,paschalidou2018raynet,ulusoy2017semantic}. 
   \item Depth map based method \cite{galliani2015massively,huang2018deepmvs,schonberger2016colmap,xu2019multigeo,yao2018mvsnet}. 
\end{inparaenum}
In this paper, we focus on the depth map based reconstruction methods and  
give a brief review of the existing learning-based works on both stereo and MVS depth estimation in this section. 

\vspace{-0.3cm}
\paragraph{Learning-based stereo.} 
Recently, learning-based works have achieved impressive results in stereo matching, 
and significantly outperform traditional stereo approaches in stereo benchmark such as KITTI \cite{geiger2012kitti}. 
Compared with handcrafted feature extraction and cost regularization, 
learning-based methods utilize the fully convolutional networks (FCNs) \cite{long2015fully} for both tasks. 
Zbontar and LeCun \cite{zbontar2015computing} train a Siamese CNN to compute the matching cost, 
which is further processed by traditional cost regularization and optimization methods to get the final depth map. 
DispNet \cite{mayer2016dispnet} adopts an end-to-end learning architecture, extends the use of CNN to the cost regularization and depth regression. 
Liang \etal \cite{liang2018iresnet} propose an iterative residual refinement sub-network to model the depth refinement. 
Unlike conventional 2-D convolutions networks,  
Kendall \etal \cite{kendall2017gcnet} apply 3-D convolutions to cost volume regularization and use differentiable depth regression to reach sub-pixel accuracy. 
Chang and Chen \cite{chang2018psmnet} improve the work of GCNet \cite{kendall2017gcnet} by incorporating the global context information into image features and 
applying stacked hourglass 3-D CNNs to the cost volume regularization. 

\vspace{-0.3cm}
\paragraph{Learning-based MVS.} 
Compared with stereo matching methods, MVS algorithms need to tackle two more issues: 
unstructured camera poses and arbitrary number of input images. 
While the first issue is usually resolved by the plane-sweep algorithm, various aggregation methods are proposed to deal with the second one in current works. 
Hartmann \etal \cite{hartmann2017patchsim} first generalize the similarity score \cite{zagoruyko2015learnpatch, zbontar2015computing,zbontar2016stereo} 
of two-view image patches to the case of multi-view by adding a mean-pooling layer for Siamese branch aggregation.  
Huang \etal \cite{huang2018deepmvs} pre-warp small image patches using the plane-sweep algorithm to build parallel unstructured two-view matching cost volumes. 
Then, these cost volumes are first passed cost volumes are first passed through a shared intra-volume regularization network and aggregated afterwards by a max pooling layer. 
A second inter-volume regularization network is applied to the aggregated intra-volume results to further refine the result. 
Since they infer depth maps patch-wisely, a dense CRF is deployed as a post-processing step for global smoothing. 
Im \etal \cite{im2019dpsnet} present DPSNet, an improved version of DeepMVS \cite{huang2018deepmvs} that 
leverages the global contextual information by extracting multi-scale deep features and 
computing matching cost volumes from the full-size image pairs rather than patches. 
Unlike DeepMVS, DPSNet uses an average-pooling layer instead to fuse the inter-volume regularization results, 
and proposes a new context-aware cost volume refinement module. 
As an alternative to the average/max pooling methods, 
Yao \etal \cite{yao2018mvsnet, yao2019rmvsnet} construct a variance based 3-D cost volume to represent the matching cost of multiple plane-sweep volumes, 
and a network is trained to infer depth map from the cost volume. 
Luo \etal \cite{Luo_2019_ICCV} propose a patch-wise matching confidence volume to increase the robustness of the cost volume in \cite{yao2018mvsnet}.
Recently, Chen \etal \cite{Chen_2019_ICCV} present a novel point-based network to refine the coarse depth prediction inferred from a variance based cost volume.

\section{Architecture overview} \label{sec:overview}
Our proposed A-TVSNet is composed of two parts: 
\begin{inparaenum}[1)]
   \item the two-view stereo network (Section~\ref{sec:basenet}) that estimates disparity\footnote{We use ``disparities'' rather than ``depths''. In the case of unstructured stereo, ``disparity'' denotes the reciprocal of depth.} 
   from an unstructured stereo image pair, and 
   \item the multi-view aggregation (Section~\ref{sec:fuse}) which is designed to fuse the multi-view information effectively. 
\end{inparaenum}

As depicted in Figure~\ref{fig:overview}, 
the two-view stereo network is divided into three distinct modules: 
the feature extraction module (FEM) (Section~\ref{sec:feature}), 
the cost regularization module (CRM) (Section~\ref{sec:cost}), 
and the refinement module (Section~\ref{sec:refine}, Figure~\ref{fig:refine}). 
The multi-view aggregation is achieved by two independent attentional aggregation modules (AAMs) 
(Section~\ref{sec:fuse}, Figure~\ref{fig:fuse}) at the end of the CRM and the refinement module respectively. 
The local information from different two-view networks is exchanged and fused 
as the global ones through AAMs to make use of the multi-view information efficiently. 

\section{Two-view stereo network} \label{sec:basenet}

In this section, we introduce a disparity
estimation and refinement network for two-view unstructured image pair. 
The two-view stereo network consists of three main modules, 
the multi-scale feature extraction module (FEM) (Section~\ref{sec:feature}), 
the cost regularization module (CRM) (Section~\ref{sec:cost}), 
and the refinement module (Section~\ref{sec:refine}). 

\subsection{Multi-scale feature extraction} \label{sec:feature} 
First, we learn a deep feature representation of input images. 
Following \cite{chang2018psmnet, zhao2017pspnet}, 
we adopt the spatial pyramid pooling (SPP) in the feature extraction to exploit both local and global contextual information. 
In this work, we use the same FEM configuration as that of PSMNet \cite{chang2018psmnet}, 
in which four fixed-size average pooling blocks $(8\times8, 16\times16, 32\times32, 64\times64)$ are used in the SPP. 
The upsampled features of different scales are then concatenated and aggregated by a $1\times1$ 2-D convolution. 
The output of the feature extraction module $\mathcal{F}^h$ is a 32-channel feature map 
and is downsized to \( \frac{1}{4} \) of the input images. 
All weights are shared among the input images. 

\subsection{Cost volume generation and regularization} \label{sec:cost} 
\paragraph{Plane-sweep cost volume.} \label{para:sweep} 
Next, a Fronto-parallel plane-sweep volume is computed using the feature maps extracted from the previous step. 
For a unstructured image pair, we define the image of which a depth map is to be estimated as the \textbf{\textit{reference image}}, 
and the other image as the \textbf{\textit{source image}}. 
The Fronto-parallel plane-sweep volume is constructed by warping the feature images onto various virtual planes. 
The virtual planes $\mathcal{D}\coloneqq\lbrace d_i \rbrace_{i=1}^{D}$ 
are the planes vertical to the Z-axis at specific distances in 
the reference image's coordinate system: 

\begin{equation} 
   d_{i}=d_{min} + i\cdot\delta 
   \label{eq:disp} 
\end{equation} 
where $D$ is the number of virtual planes ($D = 128$ in this paper), 
$d_{i}$ is the disparity value of the $i\textsuperscript{th}$ virtual plane, $d_{min}$ is the minimum disparity value 
and $\delta$ is the interval between two neighboring virtual planes.


As \cite{im2019dpsnet,kendall2017gcnet} suggested, we construct our cost volume by concatenating 
(rather than computing a distance) the two plane-sweep volumes of 
the reference feature $\mathcal{F}^h_{ref}$ and the source feature $\mathcal{F}^h_{src}$. 
Thus, a 4-D cost volume $C \in \mathbb{R}^{H\times W\times D\times 2F}$ is obtained, 
where ($H, W, F$) is the shape of the $\mathcal{F}^h$. 

\vspace{-0.5cm}
\paragraph{Cost volume regularization.} \label{para:infer} 
We then deploy a 3-D CNN to regularize the raw cost volume $C$. Inspired by \cite{chang2018psmnet,fu2019stacked}, 
our CRM consists of three stacked 3-D encoder-decoders with dense skip connections. 
The filtered cost volume $\tilde{C}$ extracted by CRM 
is followed by a output module to produce the predicted disparity map $\mathsf{\tilde{d}}$. 
The output module contains a 3-D convolution layer to reduce the number of output channel to $1$, 
and a softmax layer to obtain the disparity map's probability distribution volume $P \in \mathbb{R}^{H\times W\times D}$. 
After that, we compute the pixel-wise disparity $\mathsf{\tilde{d}}(\mathbf{u})$ as the expectation of $P$ along the depth dimension. 

For more details about the FEM and CRM architecture, please refer to the supplementary material.

\subsection{Refinement} \label{sec:refine}
\begin{figure}[t]
   \begin{center}
   \includegraphics[width=\linewidth]{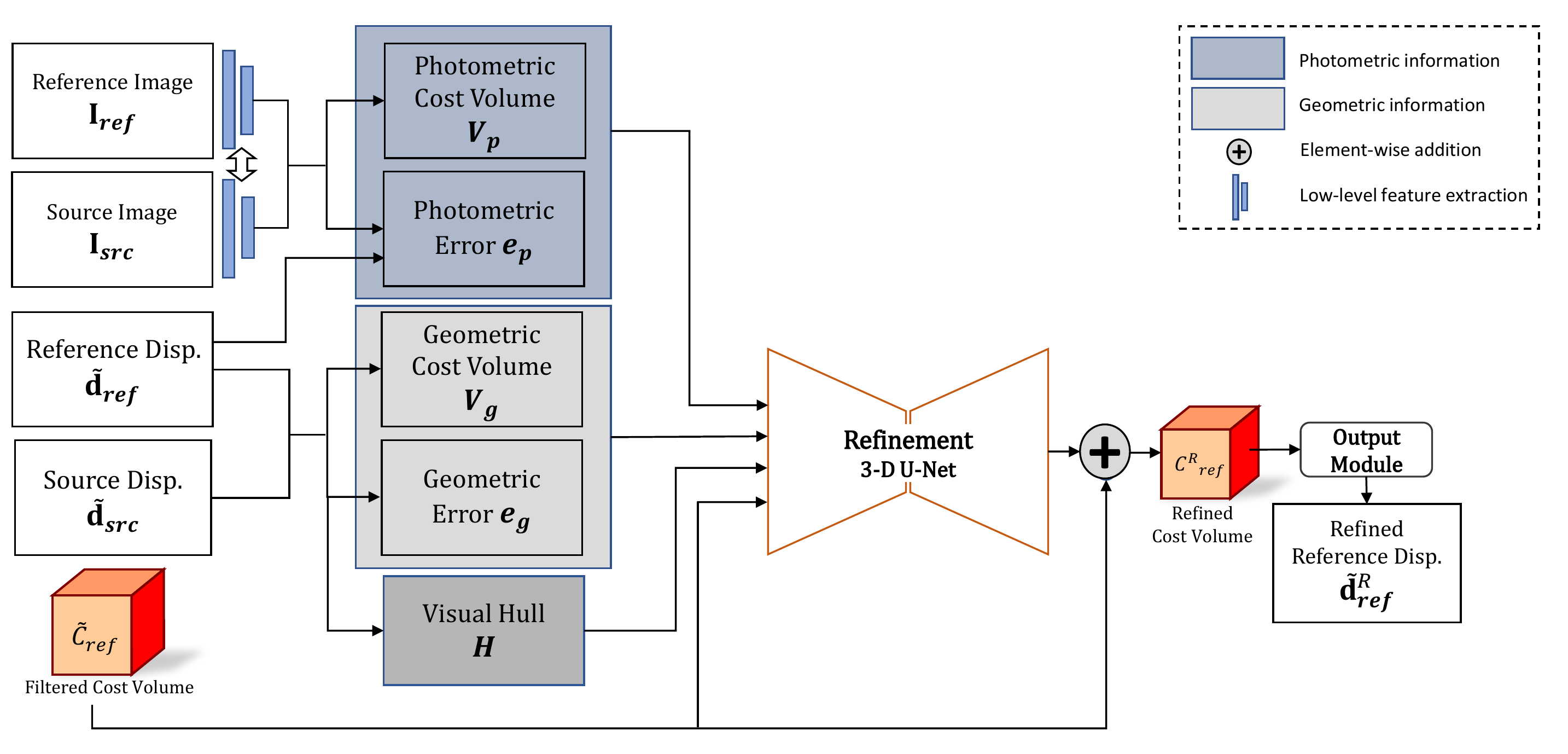}
   \end{center}
   \vspace{-0.3cm}
   \caption{The refinement architecture.}
\label{fig:refine}
\vspace{-0.2cm}
\end{figure}

To further improve the quality of our predicted disparity map, 
we introduce a residual refinement network (Figure~\ref{fig:refine}) to exploit the mutual information such as 
the photometric and geometric consistencies between the reference and source views. 
Instead of refining the output disparity value or its distribution, 
we propose to refine directly on the filtered cost volume $\tilde{C}$ 
and use the same output module as explained in Section~\ref{para:infer} to obtain the refined disparity map $\mathsf{\tilde{d}}^{R}$. 

The inputs to the refinement network are: 
the filtered cost volume $\tilde{C}$, 
the photometric and geometric consistency terms $\mathcal{V}\coloneqq\lbrace V_p,V_g \rbrace$, 
and the reconstruction errors along with the visual hull as the refinement guidance $\mathcal{G}\coloneqq\lbrace e_p,e_g,H \rbrace$.
These inputs are concatenated together and passed through a 3-D U-Net to infer the cost residual volume $C^{\Delta}$.
Then, the refined cost volume $C^{R}$ is simply the summation of $\tilde{C}$ and $C^{\Delta}$. 

Different types of information that we used for the refinement module are detailed as follows. 

\vspace{-0.3cm}
\paragraph{Photometric cost volume $V_p$}
During the refinement, we are interested in the spatial structural details (\eg, edges and boundaries) of the input images 
which can't be fully obtained from the high-level features \cite{zhao2019pyramidfeature} like those in Section~\ref{sec:feature}. 
So, we extract and use the low-level features $\mathcal{F}^l$ to construct our photometric cost volume $V_p$ 
with the method introduced in Section~\ref{sec:cost}.

\vspace{-0.3cm}
\paragraph{Geometric cost volume $V_g$} 
The geometric cost volume $V_g$ is defined in the frustum volume of reference camera. 
Given the two estimated disparity maps $\mathsf{\tilde{d}}_{ref}$ and $\mathsf{\tilde{d}}_{src}$, 
we define the geometric cost volume of the source view $V_g^{src}$ at location $(\mathbf{u}, d_i)$ as: 
\begin{equation} 
   V_g^{src} \left( \mathbf{u}, d_i \right) = \left| \mathsf{\tilde{d}}_{src}^{*} \left( \pi_{src} \left( \mathbf{u}, \mathsf{\tilde{d}}_{ref} \left( \mathbf{u} \right) \right) \right) - d_i \right| 
   \label{eq:geo_cost}
\end{equation} 
where $\mathbf{u}$ is the reference pixel coordinate vector, 
$\pi_{src}(\cdot)$ is the function that projects $\mathbf{u}$ onto the corresponding source pixel coordinate 
with given reference disparity value $\mathsf{\tilde{d}}_{ref}(\mathbf{u})$. 
$\mathsf{\tilde{d}}_{src}^{*}$ denotes the rescaled disparity map of $\mathsf{\tilde{d}}_{src}$ 
in the reference coordinate system: 
\begin{equation} 
   \mathsf{\tilde{d}}_{src}^{*}(\mathbf{u})=\frac{\mathsf{\tilde{d}}_{src}(\mathbf{u})}{ \left[ \mathtt{P}_{ref}\mathtt{P}_{src}^{-1} 
   (\mathbf{u}^\mathsf{T},1)^\mathsf{T} \right]_Z} 
   \label{eq:trans_disp} 
\end{equation} 
with $\mathtt{P}_{ref}$/$\mathtt{P}_{src}$ the projection matrix of the reference/source camera, 
and $[\cdot]_Z$ the Z component of a coordinate vector. 
Without loss of generality, the geometric cost volume for the reference disparity map is 
the distance between the estimated disparity and the disparity hypothesis: 
\begin{equation} 
   V_g^{ref} \left( \mathbf{u}, d_i \right) = \left| \mathsf{\tilde{d}}_{ref}(\mathbf{u}) - d_i \right|\text{.} 
\label{eq:geo_cost_ref} 
\end{equation} 
Then we concatenate $V_g^{ref}$ and $V_g^{src}$ to obtain the geometric cost volume $V_g$. 

While the initial estimates are inferred solely from the photometric information, 
$V_g$ acts as an additional geometric clue in the refinement module to enforce 
geometric consistency between the predicted reference and source disparity maps. 


\vspace{-0.3cm} 
\paragraph{Photometric error $e_p$} 
Given a pair of low-level image features $\mathcal{F}^l$, and $\mathsf{\tilde{d}}_{ref}$. 
The photometric error $e_p$ is computed as below: 
\begin{equation} 
   e_{p}(\mathbf{u}) = \left| \mathcal{F}^l_{src} \left( \pi_{src} \left( \mathbf{u}, \mathsf{\tilde{d}}_{ref} \left( \mathbf{u} \right) \right) \right) - \mathcal{F}^l_{ref}(\mathbf{u}) \right|
   \label{eq:photo_err}
\end{equation}
with $\mathcal{F}^l(\mathbf{u})$ the pixel value of feature map $\mathcal{F}^l$ at the pixel coordinate $\mathbf{u}$. 

\vspace{-0.3cm}
\paragraph{Geometric error $e_g$} 
Given $\mathsf{\tilde{d}}_{src}$ and $\mathsf{\tilde{d}}_{ref}$, the geometric error $e_g$ is defined as: 
\begin{equation} 
   e_{g}(\mathbf{u}) = \left| \mathsf{\tilde{d}}_{src}^{*} \left( \pi_{src} \left( \mathbf{u}, \mathsf{\tilde{d}}_{ref} \left( \mathbf{u} \right) \right) \right) - \mathsf{\tilde{d}}_{ref}(\mathbf{u}) \right|.
   \label{eq:geo_err}
\end{equation} 

We repeat $e_p$ and $e_g$ for $D$ times along the depth dimension to fit the 3-D refinement network. 
$e_p$ and $e_g$ are the reconstruction errors that measure the reliability of the initial estimations and 
provide a guided mask indicates whether a pixel's disparity needs to be further refined.

\vspace{-0.3cm} 
\paragraph{Visual hull $H$} 
Visual hull \cite{kutulakos2000carving,laurentini1994visualhull} is the intersection of multiple back-projected generalized cones that defined by the silhouette masks. 
In practice, we project each voxel $(\mathbf{u}, d_i)$ in the reference frustum space to each of the input views and set to 1 (visible) 
if the projected voxel is not occluded by that view’s estimated disparity map, 
otherwise, set it to 0 (occluded). 
The visibility degrees from different views are then summed up and normalized to form the visual hull $H$, which is formulated as: 
\begin{equation} 
   H(\mathbf{u}, d_i) = \frac{1}{N} \sum_{n=1}^{N} {\bm\theta} \left( \mathsf{\tilde{d}}_{n} \left( \pi_{n} \left( d_i \left( \mathbf{u} \right), \mathbf{u} \right) \right) - d_i \right)
   \label{eq:vis_hull}
\end{equation}
where $\bm\theta(\cdot)$ is the unit step function, $N$ is the number of input images ($N=2$ for the two-view network). 
As inconsistency often happens on surface points or noisy area, 
$H$ can provide valuable information on the global consistency in 3D space and 
thus guides the refinement module to make further improvements. 

\begin{figure} 
   \begin{center}
   \includegraphics[width=\linewidth]{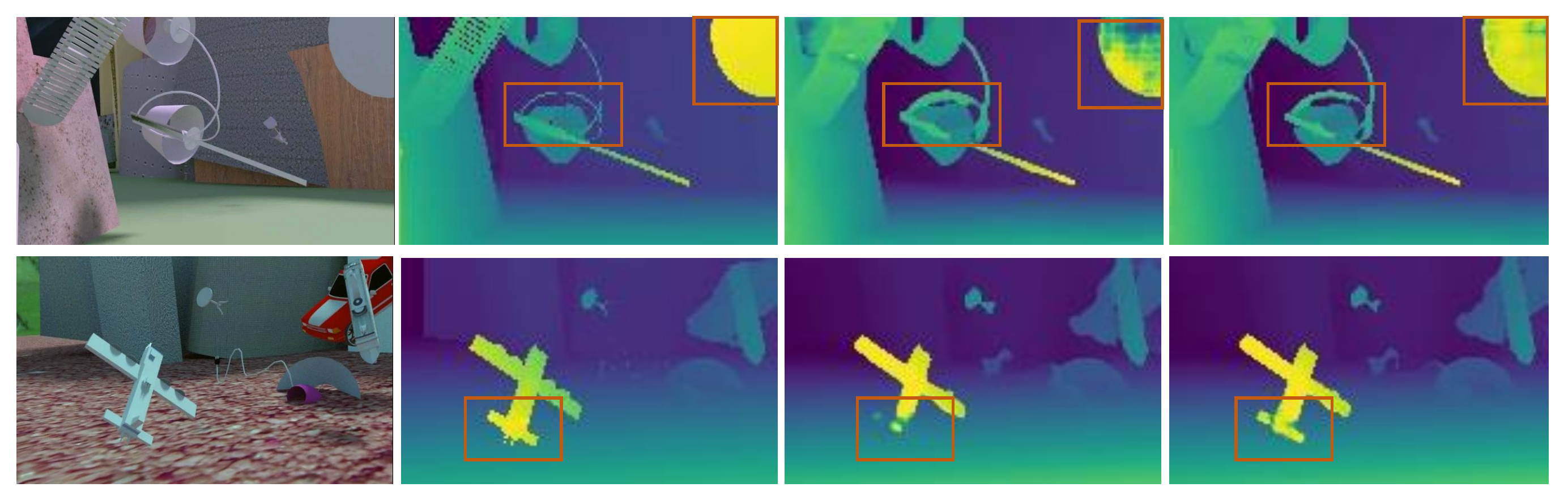}
   \end{center}
   \vspace{-0.4cm}
   \caption{Disparity refinement results. From left: reference image, ground truth disparity map, initial disparity map, refined disparity map. 
      }
\label{fig:comparerefine}
\vspace{-0.2cm}
\end{figure}

Some results of the refinement network are shown in Figure~\ref{fig:comparerefine}, 
we can see that it indeed recovers the missing details of the initial estimates from the base network. 

\subsection{Loss functions} \label{sec:loss} 
We use the mean absolute error between the estimated disparity map $\mathsf{\tilde{d}}$ and 
the ground truth disparity map $\mathsf{{d}}^{\prime}$ 
as our loss function $\ell ( \mathsf{\tilde{d}},\mathsf{{d}}^{\prime} )$. 
Unavailable or invalid pixels in the ground truth disparity map are ignored. 
Intermediate supervisions similar to \cite{chang2018psmnet} are applied to facilitate the training process, 
with the training loss defined as: 
\begin{equation} 
   L = \lambda \ell ( \mathsf{\tilde{d}}^{R},\mathsf{{d}}^{\prime} ) + \sum_{k=1}^{3} \omega^{k} \ell ( \mathsf{\tilde{d}}^{k},\mathsf{{d}}^{\prime} )
   \label{eq:loss_total}
\end{equation}
where $\lambda$ is the weight of the refined disparity map, 
$\mathsf{\tilde{d}}^{k}$ is the disparity map 
produced by the $k\textsuperscript{th}$ encoder-decoder of CRM 
and $\omega^{k}$ is the corresponding weight.

\section{Multi-view aggregation} \label{sec:fuse} 
In this section, we will detail our strategy for aggregating multi-view information. 

As mentioned in Section~\ref{sec:intro}, we argue that a good aggregation module 
should enable efficient information exchange and intergration among different views. 
To this end, following the idea from \cite{aittala2018burst}, 
we propose a multi-view aggregation framework that allows multiple aggregation operations at flexible locations. 
At each aggregating point, the local information flows from different two-view networks are exchanged in the form of the global information, 
and both of the local and the global information are utilized throughout the network. 

In A-TVSNet, we propose to use two aggregation operations (AAM1 and AAM2 as shown in Figure~\ref{fig:overview}). 
The first aggregation operation (AAM1) takes place right after the CRM: 
it intergrates the $N$ filtered cost volumes of the reference image 
${\lbrace\tilde{C}_{ref}^n\rbrace_{n=1}^{N}}$ into a single filtered cost volume $\mbox{\small $\hat{C}_{ref}$ }$. 
AAM1 enables a first information exchange between the $N$ two-view networks and 
forces them to have the same estimate on the reference initial disparity map $\mathsf{\tilde{d}}_{ref}$. 
Meanwhile, estimates of the source initial disparity map $\mathsf{\tilde{d}}_{src}$ 
as well as the low-level feature maps $\mathcal{F}^l_{src}$ are kept as 
each network's own local information for further processings in the refinement module. 
The second aggregation operation (AAM2) happens right before the last output module: 
it intergrates the $N$ refined cost volumes of the reference image 
$\lbrace {C^{R}}_{ref}^n \rbrace_{n=1}^{N}$ into a single refined cost volume $\mbox{\small $\widehat{C^{R}}_{ref}$ }$. 
In contrast to AAM1, AAM2 aggregates all the local knowledge of the $N$ two-view networks 
and none of the information is kept locally afterwards. 
Thus, the $N$ two-view networks become the same after AAM2, 
and the final estimate of the reference disparity map $\mathsf{\tilde{d}}^{R}_{ref}$ 
is obtained by passing $\mbox{\small $\widehat{C^{R}}_{ref}$ }$ to the output module. 

It is noteworthy that our aggregation framework does not restrict to any sepsific aggregation module. 
Exploring for the optimal configuration of aggregation framework remains as future work.

\begin{figure}[t]
   \begin{center}
   \includegraphics[width=0.9\linewidth]{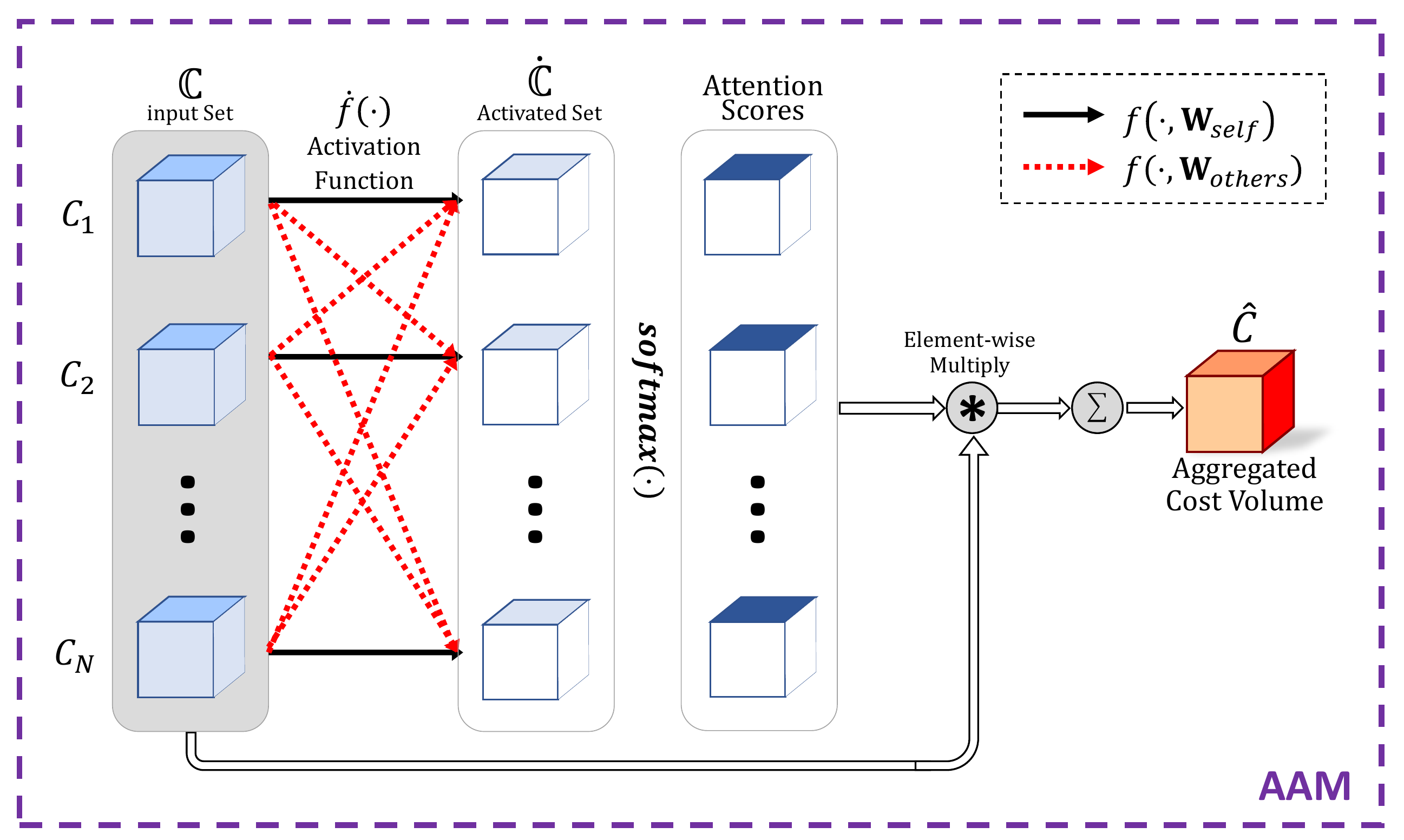}
   \end{center}
   \vspace{-0.3cm}
   \caption{The figure illustrates our attentional aggregation module (AAM). 
      In contrast to \cite{Yang2018AttentionalAO}, we take the mutual information into consideration 
      by introducing the second activation (red dash line connections in figure).} 
\label{fig:fuse}
\vspace{-0.2cm}
\end{figure}

\begin{figure} 
   \begin{center}
   \includegraphics[width=0.9\linewidth]{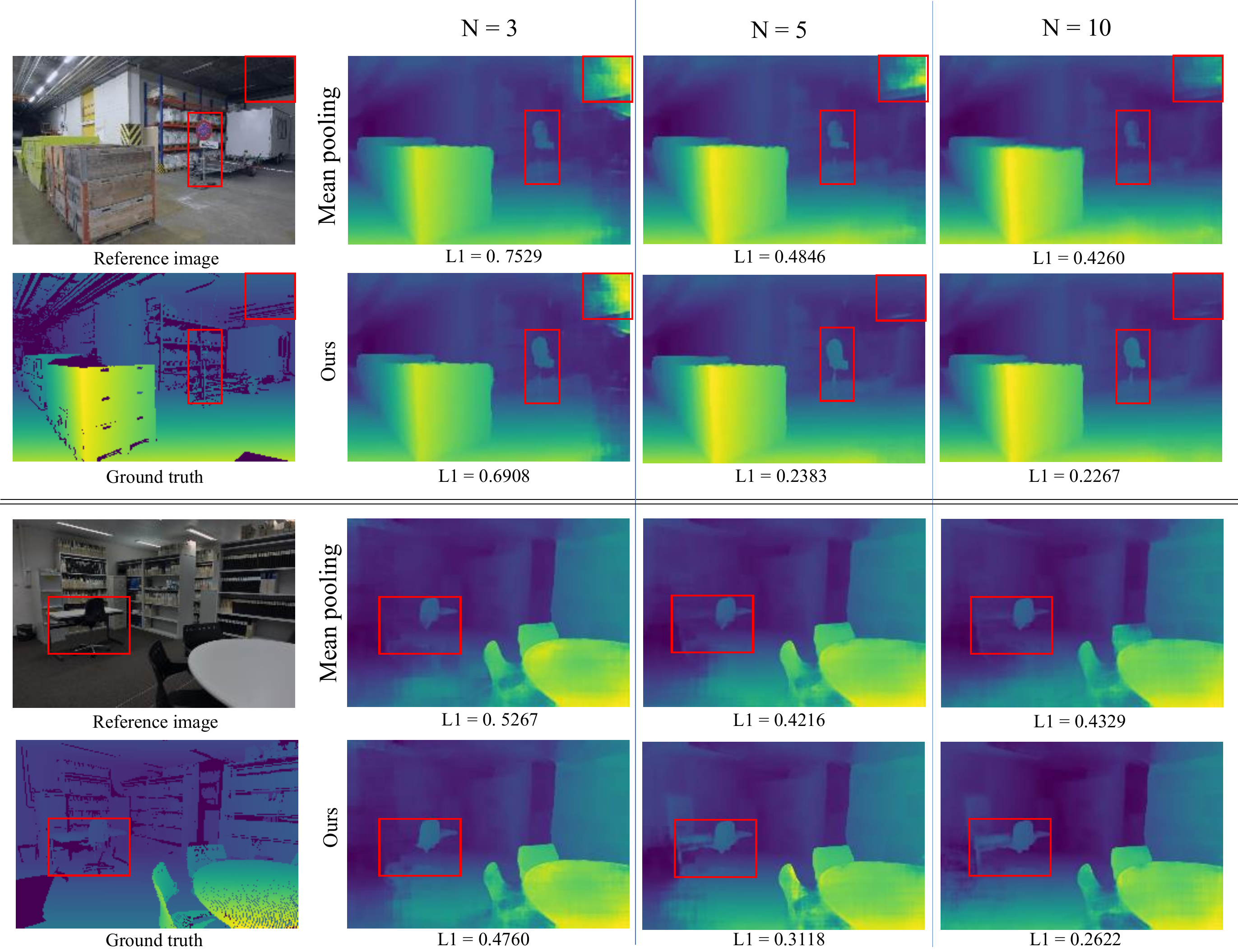}
   \end{center}
   \vspace{-0.5cm}
      \caption{The figure shows comparisons between the disparity maps aggregated by mean pooling and our AAM. 
      }
\label{fig:comparefuse}
\vspace{-0.2cm}
\end{figure}

\vspace{-0.3cm}
\paragraph{Attentional aggregation module.} 
As for the specific design of aggregation modules, 
the mainstream solutions are flawed in different aspects. 
RNNs based strategies are order variant, 
and put different frames into a highly asymmetric position \cite{aittala2018burst}. 
Pooling operations are too simple to retain enough underlying information from all cost volumes, 
and thus the improvement is often limited (Figure~\ref{fig:comparefuse}). 

To alleviate these problems, Yang \etal \cite{Yang2018AttentionalAO} introduce AttSets, 
a permutation invariant aggregation module with attentional scores. 
The aggregation module learns attention scores by a non-linear activation function 
for all elements within the input set. 
These scores can be regarded as an attentional mask that helps to select useful features, 
which are then aggregated by a weighted averaging operation. 

Our proposed attentional aggregation module (AAM) extends AttSets \cite{Yang2018AttentionalAO} 
to leverage the mutual information of multiple cost volumes (Figure~\ref{fig:fuse}). 
In AttSets, each attention score is solely determined by the corresponding cost volume and 
the inter information of other cost volumes is ignored. 
We add a second shared weights non-linear activation function 
to model the interrelationship between the corresponding cost volume and the others. 

Specifically, given a set of cost volumes $\mathbb{C}\coloneqq\lbrace C_{n} \rbrace_{n=1}^N$ as input, 
the $n\textsuperscript{th}$ element of the activated set $\dot{\mathbb{C}}\coloneqq\lbrace \dot {C}_{n} \rbrace_{n=1}^N$ is calculated as: 
\begin{equation} 
\begin{aligned}
   \dot{C}_{n}&=\dot{f}(\mathbb{C}, \mathbf{W}_{self}, \mathbf{W}_{others}) \\
   &=f(C_{n},\mathbf{W}_{self})+\sum_{m=1\atop m\neq n}^{N} f(C_{m},\mathbf{W}_{others}) 
\end{aligned}
\label{eq:fuse_activation}
\end{equation} 
where $f(\cdot)=conv3D(\cdot)$ is the attention activation function as in AttSets. 
$\mathbf{W}_{self}$ and $\mathbf{W}_{others}$ in the activation function are learnable parameters for  
the corresponding cost volume and the rest cost volumes respectively.

Then, the aggregated cost volume $\hat{C}$ is computed as the weighted sum of $\mathbb{C}$:  
\begin{equation} 
\begin{aligned}
   \hat{C}=\sum_{n=1}^{N} C_{n}\cdot softmax(\dot{\mathbb{C}})_{n}.
\end{aligned}
\label{eq:fuse_sum}
\end{equation} 

As shown in Figure~\ref{fig:comparefuse} and Table~\ref{tab:ablation_fuse}, our AAM is more robust and effective 
than the pooling method especially when the number of input views increases,   
and the second activation function also yields moderate performance improvement over AttSets. 
More comparative results are available in the ablation study (Section~\ref{sec:ablation_fuse}). 

\begin{figure*}[t]
   \begin{center}
   \subfloat[DeMoN testset qualitative results (2 views).\label{fig:comparealgo2}]{\includegraphics[width = 0.95\linewidth]{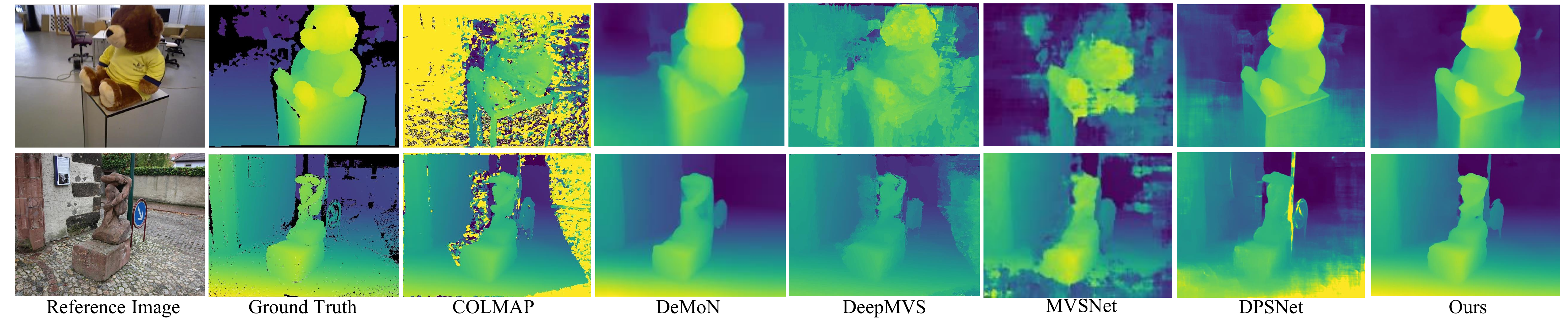}}
   \\ 
   \subfloat[ETH3D dataset qualitative results (5 views).\label{fig:comparealgo5}]{\includegraphics[width = 0.95\linewidth]{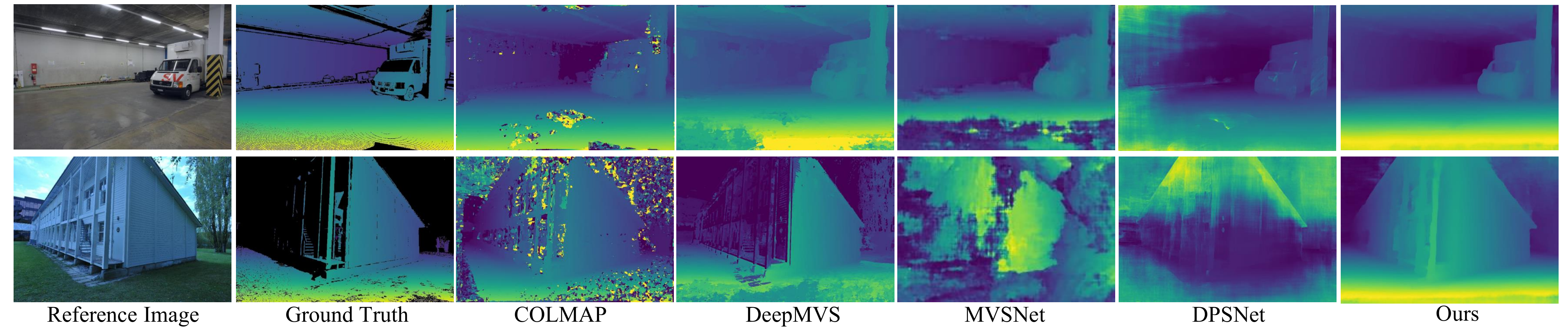}}
   \end{center}
   \vspace{-0.5cm}
   \caption{Qualitative comparison for disparity estimates between different algorithms. 
   From left, the reference image, ground truth, COLMAP \cite{schonberger2016colmap}, DeMoN \cite{UZUMIDB17demon}, DeepMVS \cite{huang2018deepmvs}, 
   DPSNet \cite{im2019dpsnet}, MVSNet \cite{yao2018mvsnet} and ours. 
   (a) is the result for two-view input on DeMoN testset and 
   (b) is the result for multi-view (use 5 views in figure) input on ETH3D dataset. } 
   \label{fig:comparealgo}
   \vspace{-0.2cm}
\end{figure*}

\section{Experiments}\label{sec:result}

\subsection{Datasets}\label{sec:result_dataset}
Our training data consists of two real-world datasets: \textbf{Achteck-Turm} and \textbf{Citywall} \cite{fuhrmann2015mve}, \textbf{SUN3D} \cite{xiao2013sun3d};  
and two synthesized datasets: \textbf{Scenes11} \cite{chang2015shapenet,UZUMIDB17demon}, \textbf{MVS-Synth} \cite{huang2018deepmvs}. 
Each dataset contains a number of short sequences of images with corresponding camera parameters and ground truth depth maps. 

We use DeMoN \cite{UZUMIDB17demon} testset to evaluate the performance of the two-view stereo network ($N$=2). 
For multi-view stereo evaluation ($N$\textgreater2), we use the multi-view dataset of ETH3D \cite{schoeps2017eth3d} 
which contains sequences of real-world images with ground truth point clouds. 
The ground truth point clouds are back-projected to get corresponding disparity maps. 

\subsection{Implementation details} \label{sec:result_detail}
A-TVSNet is implemented with TensorFlow \cite{abadi2016tensorflow} and trained on one Nvidia 1080Ti graphics card. 
We train our model using a two-stage training strategy. 
First, the two-view stereo network (Section~\ref{sec:basenet}) is trained from scratch for 1000K iterations, 
with weights $\lambda=0.8$, $\omega^{1}=0.2$, $\omega^{2}=0.3$ and $\omega^{3}=0.5$ in Equation~\ref{eq:loss_total}. 
Second, the two aggregation modules (Section~\ref{sec:fuse}) are trained together for 200K iterations with the number of input views $N=3$. 
The weights of the two-view stereo network are frozen in this stage, 
and the intermediate losses are disabled. 
During each training stage, the RMSProp optimizer \cite{Tieleman2012rmsprop} is applied with mini-batch size of 16, 
the initial learning rate is 0.001 which is decreased by $0.9$ for every 10K iterations. 

\begin{table} [H]
   \begin{center}
      \resizebox{\linewidth}{!}{\begin{tabular}{@{\extracolsep{5pt}}l| cccc cccc@{}}
      \hline
      \multirow{2}{*}{Method} & \multicolumn{4}{c}{Error (less is better)} & \multicolumn{4}{c}{Accuracy$(\%)$ (larger is better)}
      \\ \cline{2-5} \cline{6-9}
      & L1 & L1-inv & L1-rel & Sc-inv & $<\delta$ & $<3\delta$ & $<5\delta$ & $<10\delta\:\:$\\
      \hline\hline
      \multicolumn{9}{c}{DeMoN (2 views)}\\
      \hline
      COLMAP                               & 5.5854      & 6.1850       & 1.0606       & 1.0261       & 36.62        & 44.23        & 48.64        & 56.32\\
      DeMoN                                & 8.9631      & \bf{0.0300}  & 0.2418       & 0.2017       & 23.63        & 41.83        & 52.53        & 66.15\\
      DeepMVS                              & 2.8724      & 0.0877       & 0.2605       & 0.3501       & 38.62        & 52.99        & 59.74        & 69.54\\
      DPSNet                               & 2.0774      & 0.0651       & 0.1246       & 0.2414       & 49.63        & 64.09        & 71.10        & 80.86\\
      MVSNet                               & 5.7666      & 0.1022       & 0.8759       & 0.4576       & 32.38        & 45.54        & 52.95        & 64.43\\
      Ours w/o refine                      & 2.0039      & 0.0373       & 0.0995       & 0.2030       & 49.38        & 64.19        & 71.75        & 82.32\\
      Ours                                 & \bf{1.9290} & 0.0357       & \bf{0.0949}  & \bf{0.1957}  & \bf{49.85}   & \bf{64.76}   & \bf{72.38}   & \bf{82.88}\\
      \hline\hline
      \multicolumn{9}{c}{ETH3D (5 views)}\\
      \hline
      COLMAP                               & 0.9734      & 0.0380       & 0.2990       & 0.4697       & \bf{68.67}   & 75.58        & 78.27        & 82.11\\
      DeepMVS                              & 1.2456      & 0.0479       & 0.3170       & 0.2553       & 48.06        & 68.32        & 74.94        & 82.14\\
      DPSNet                               & 1.7160      & 0.0843       & 0.1896       & 0.3149       & 30.76        & 51.87        & 61.50        & 73.52\\
      MVSNet                               & 3.6419      & 0.0923       & 0.9736       & 0.4820       & 38.53        & 53.70        & 58.94        & 66.07\\
      Ours w/o refine                      & 0.4964      & 0.0343       & 0.1188       & 0.1586       & 57.03        & 75.13        & 81.44        & 88.36\\
      Ours                                 & \bf{0.4763} & \bf{0.0329}  & \bf{0.1154}  & \bf{0.1573}  & 58.77        & \bf{76.87}   & \bf{82.82}   & \bf{89.17}\\
      \hline
      \end{tabular}}
   \end{center}
   \vspace{-0.2cm}
   \caption{Quantitative comparisons between different MVS algorithms on DeMoN testset and ETH3D dataset.}
   \label{tab:eval_algo}
   \vspace{-0.1cm}
\end{table}

\subsection{Evaluations} \label{sec:result_eval_depth}
\paragraph{Depth map evaluation.}
We compare the depth estimation results of A-TVSNet with several state-of-the-art MVS algorithms, 
including a conventional method COLMAP \cite{schonberger2016colmap} 
and learning-based methods, \ie, DeMoN \cite{UZUMIDB17demon} (only works with two-view image pairs), 
DeepMVS \cite{huang2018deepmvs}, MVSNet \cite{yao2018mvsnet}, DPSNet \cite{im2019dpsnet}. 
These algorithms are evaluated on DeMoN two-view testset and ETH3D multi-view dataset. 
DeMoN testset consists of sequences of two-view unstructured image pairs from multiple datasets 
(\ie, MVS \cite{fuhrmann2015mve}, SUN3D \cite{xiao2013sun3d}, Scenes11 \cite{chang2015shapenet}, RGBD \cite{sturm12rgbd}). 
ETH3D dataset is used to evaluate multi-view performance, and we take five images as input in this experiment. 
All input images are resized to $960\times640$, 
and neither post-processing nor filtering is applied during depth map evaluations. 

Figure~\ref{fig:comparealgo} shows qualitative disparity map comparsions between our A-TVSNet and other algorithms.  
It can be seen that the disparity maps produced by our network have fewer noisy predictions and artifacts on low-textured areas. 
In order to measure the performance of our network quantitatively, we use mean absolute error (L1), inverse mean absolute error (L1-inv), 
relative mean absolute error (L1-rel) as well as scale-invariant error \cite{eigen2014depth} (Sc-inv) as error metrics. 
For accuracy metric, we adopt inlier ratio (\textless$k\delta$), $k\!\!\:\in\!\!\:\lbrace1,3,5,10\rbrace$  
which indicates the percentage of pixels whose errors are below a certain threshold. 
As demonstrated in Table~\ref{tab:eval_algo}, 
our network shows better performance on both two-view and multi-view datasets. 

\vspace{-0.3cm}
\paragraph{Point cloud evaluation.}
We generate point clouds from all estimated depth maps using the similar filtering and fusion method provided by \cite{galliani2015massively}. 
Table~\ref{tab:eval_pc} shows quantitative results of point cloud reconstruction, 
more details and comparsions of point cloud evaluation are shown in the supplementary material. 

\begin{table} [h]
   \begin{center}
      \resizebox{\linewidth}{!}{\begin{tabular}{c | ccc }
      \hline
      Tolerance(cm) & $F_1$ Score $(\%)$ & Accuracy $(\%)$ & Completeness $(\%)\:$\\
      \hline
      2           & 42.12   & 33.17      & 61.02\\
      5           & 63.67   & 55.66      & 75.66\\
      10          & 77.14   & 72.66      & 82.55\\
      \hline
      \end{tabular}}
   \end{center}
   \vspace{-0.3cm}
   \caption{Point cloud evaluation results of A-TVSNet on ETH3D benchmark\protect\footnotemark.}
   \label{tab:eval_pc}
   \vspace{-0.2cm}
\end{table}
\footnotetext{\url{https://www.eth3d.net/low_res_many_view}}

\vspace{-0.2cm}
\subsection{Ablation studies} \label{sec:ablation}
In this section, two ablation studies are analyzed to justify the efficacy of our network designs.

\vspace{-0.3cm}
\paragraph{Refinement.} \label{sec:ablation_refine}
To quantify the contributions of different types of information in our refinement network,
we retrain two refinement networks without $\lbrace V_g, e_g, H \rbrace$ and without $\lbrace H \rbrace$ respectively. 
Table~\ref{tab:ablation_refine} shows that the photometric terms $\lbrace V_p, e_p \rbrace$, 
the geometric terms $\lbrace V_g, e_g \rbrace$ and the visual hull term $\lbrace H \rbrace$ each can provide improvements in error metrics. 

\begin{figure} [h]
   \begin{center}
   \includegraphics[width=\linewidth]{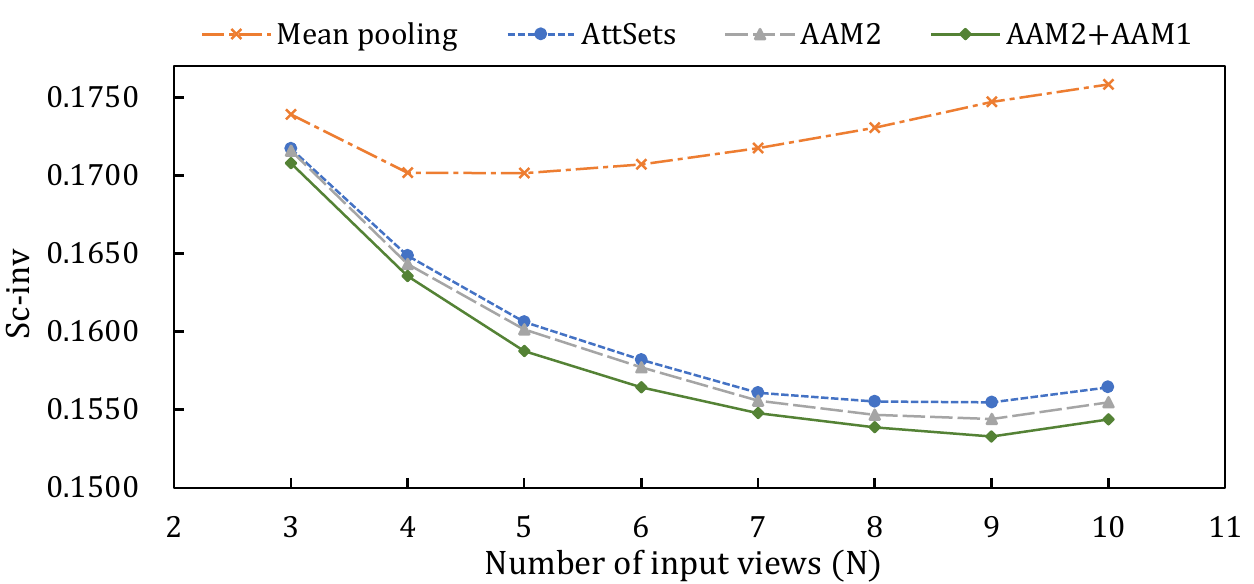}
   \end{center}
   \vspace{-0.3cm}
   \caption{The figure shows the tendency of Sc-inv error for increasing number of input views on ETH3D dataset with different aggregation methods. }
\label{fig:ablation_fuse_scinv}
\vspace{-0.2cm}
\end{figure}

\vspace{-0.3cm}
\paragraph{Aggregation module.} \label{sec:ablation_fuse}
In this part, we quantitatively compare three aggregation methods with A-TVSNet. 
In all these three methods, we keep only one aggregation operation at the location of AAM2. 
Different aggregation modules, \ie, mean pooling, AttSets and AAM are tested. 
As shown in Table~\ref{tab:ablation_fuse} and Figure~\ref{fig:ablation_fuse_scinv}, 
the pooling aggregation does not always improve the results as the number of input images increases, 
because weights are equal for each view and thus bad estimates introduced by large baseline pairs, occlusions, \etc, will impair the aggregated result.  
It can also be observed that our AAM outperforms AttSets with the help of the second activation. 
Moreover, adding another aggregation module (AAM1) improves the overall depth estimation performance. 

\begin{table} [h]
   \begin{center}
      \resizebox{\linewidth}{!}{\begin{tabular}{l|cccc}
      \hline
      \multirow{2}{*}{Refinement architecture} & \multicolumn{4}{c}{Error}
      \\ \cline{2-5}
      & L1 & L1-inv & L1-rel & Sc-inv
      \\
      \hline\hline
         Without refinement                                       & 2.0039       & 0.0373       & 0.995        & 0.2030\\
      \hline\hline
         Refine w/o $\lbrace V_g, e_g, H \rbrace$                 & 1.9685       & 0.0367       & 0.0974       & 0.2012\\
         Refine w/o $\lbrace H \rbrace$                           & 1.9321       & 0.0364       & \bf{0.0949}  & 0.1976\\
         Full refinement                                          & \bf{1.9209}  & \bf{0.0357}  & \bf{0.0949}  & \bf{0.1957}\\
      \hline
      \end{tabular}}
   \end{center}
   \vspace{-0.3cm}
   \caption{Quantitative comparisons between different refinement architectures on DeMoN testset (2 views).}
   \label{tab:ablation_refine}
   \vspace{-0.2cm}
\end{table}

\begin{table} [h]
   \begin{center}
   \resizebox{\linewidth}{!}{\begin{tabular}{l| ccc | ccc | ccc }
   \hline
   \multirow{2}{*}{Method} & \multicolumn{3}{c|}{L1} & \multicolumn{3}{c|}{L1-inv} & \multicolumn{3}{c}{Sc-inv} \\ \cline{2-10}
                           & N=3 & N=5 & N=10 & N=3 & N=5 & N=10 & N=3 & N=5 & N=10 \\
   \hline\hline
   Mean pooling                           & 0.5929          & 0.5860          & 0.6458          & 0.0390          & 0.0405          & 0.0444          & 0.1739          & 0.1701          & 0.1758             \\
   AttSets \cite{Yang2018AttentionalAO}   & 0.5571          & 0.4978          & 0.4811          & 0.0357          & 0.0336          & 0.0333          & 0.1717          & 0.1606          & 0.1564             \\
   AAM2                                   & 0.5589          & 0.4963          & 0.4733          & 0.0357          & 0.0334          & 0.0329          & 0.1715          & 0.1601          & 0.1555             \\
   A-TVSNet (AAM2+AAM1)                    & \textbf{0.5501} & \textbf{0.4736} & \textbf{0.4677} & \textbf{0.0355} & \textbf{0.0329} & \textbf{0.0324} & \textbf{0.1707} & \textbf{0.1573} & \textbf{0.1543}   \\  
   \hline
   \end{tabular}}
   \end{center}
   \vspace{-0.3cm}
   \caption{Quantitative comparisons between different aggregation methods on ETH3D dataset. }
   \label{tab:ablation_fuse}
   \vspace{-0.2cm}
\end{table}

\section{Conclusion}
We have developed an effective learning-based network A-TVSNet for depth estimation from multi-view stereo images. 
In A-TVSNet, the MVS network is reformulated into multiple two-view stereo networks with information communication and aggregation. 
We propose a permutation-invariant aggregation framework to efficiently exchange and integrate multi-view information. 
Furthermore, our refinement network is able to exploit geometric information to produce high quality depth maps. 
Finally, our MVS system shows better depth map reconstruction quality 
than competing MVS approaches in challenging indoor and outdoor scenes. 


{\small
\bibliographystyle{ieee_fullname}
\bibliography{egbib}
}

\newpage
\section{Supplementary Material}

\begin{figure*}
   \begin{center}
   \includegraphics[width=0.98\linewidth]{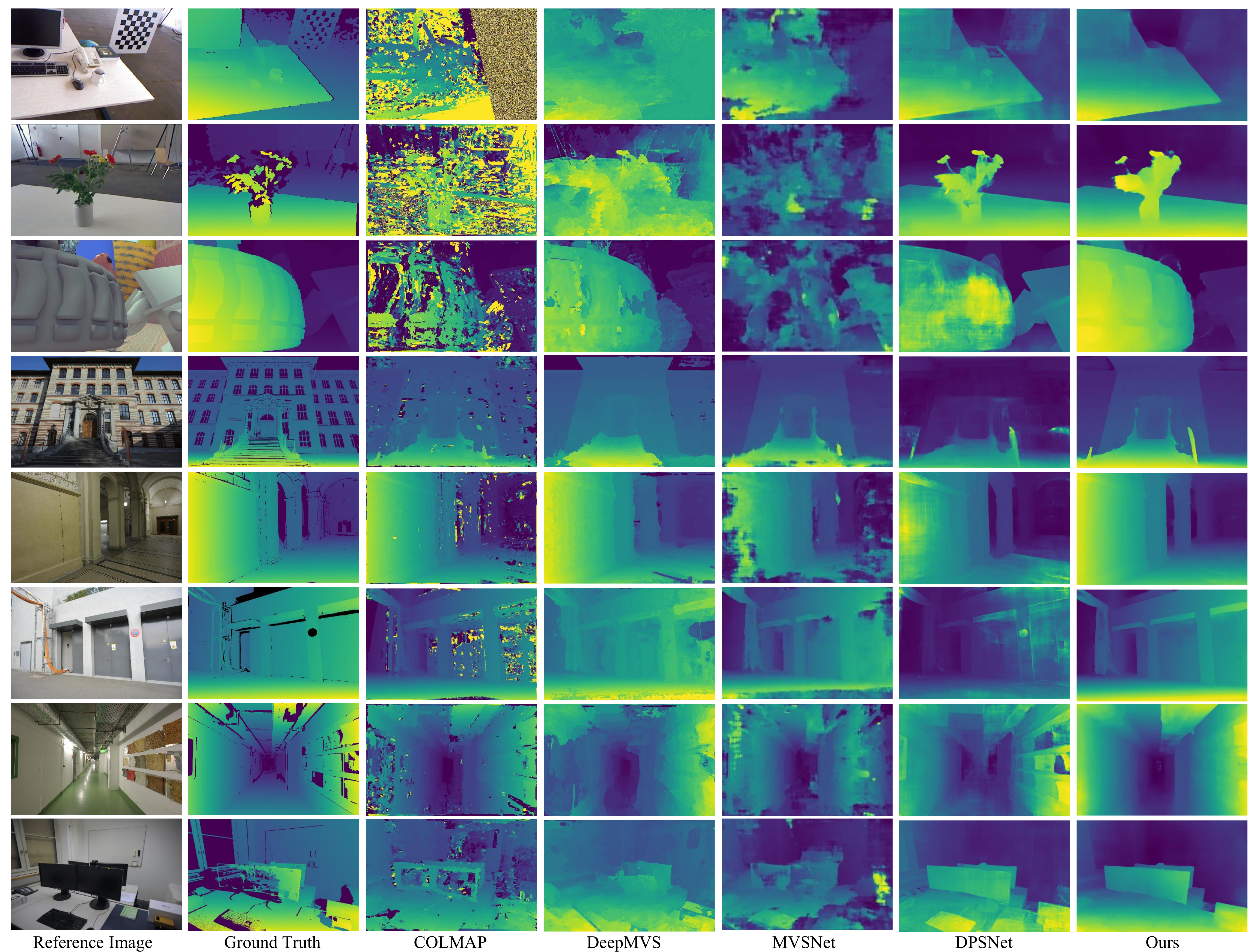}
   \end{center}
   \caption{More qualitative comparison for disparity estimation between different algorithms. 
   First three rows are from the DeMoN \cite{UZUMIDB17demon} dataset, others are from the ETH3D dataset.}
   \label{fig:algo_comp_more}
\end{figure*}

In this supplementary material, we first describe the detailed network architecture and additional implementation details. 
We then show more point cloud evaluation results to complement the main paper, 
more qualitative disparity estimation results are available in Figure~\ref{fig:algo_comp_more}.

\subsection{Network architecture}
\begin{figure} [htb]
   \begin{center}
   \includegraphics[width=\linewidth]{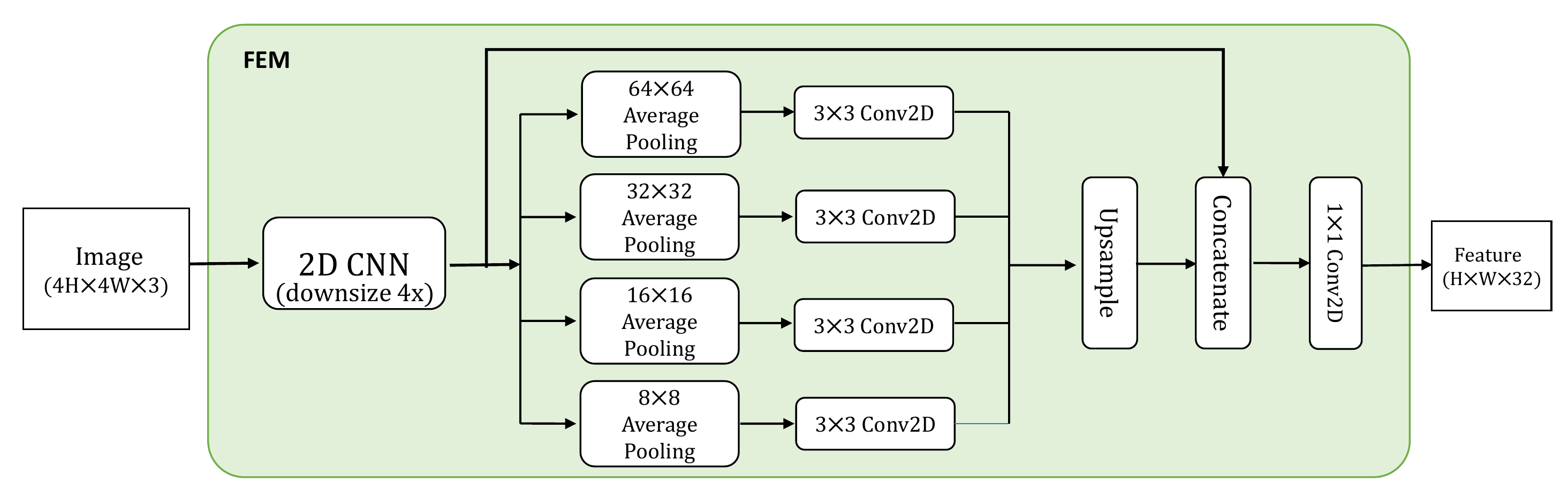}
   \end{center}
      \caption{The detailed design of feature extraction module (FEM).
      }
\label{fig:fem}
\end{figure}

The feature extraction module (FEM) as shown in Figure~\ref{fig:fem} is similar to that of PSMNet \cite{chang2018psmnet} which consists of a 2-D CNN and a SPP module.
The 2-D CNN contains three $3\times3$ convolution and four cascaded residual blocks \cite{he2016residual}, 
the strides of the first convolution and the first residual block are set to 2 to make the output feature map size \( \frac{1}{4} \) of the input image size.

\begin{figure} [htb]
   \begin{center}
   \includegraphics[width=\linewidth]{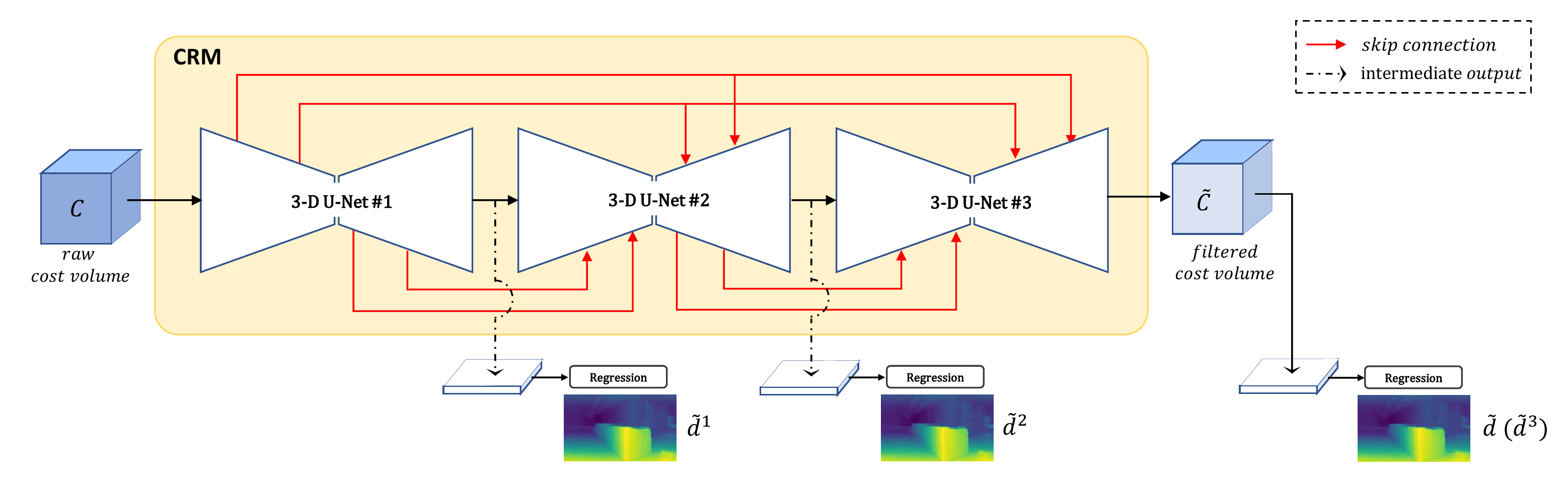}
   \end{center}
      \caption{The detailed design of cost regularization module (CRM).}
\label{fig:crm}
\end{figure}

The cost regularization module (CRM) mainly consists of three stacked U-Net like 3-D CNNs. 
Each individual 3-D CNN has the same structure as in \cite{yao2018mvsnet}, 
and the corresponding depth map can be obtained through an additional output module, 
skip connections are also added between different U-Net structures as Figure~\ref{fig:crm} shows.

\subsection{Point cloud evaluation}

\begin{table*} 
   \begin{center}
      \resizebox{0.98\linewidth}{!}{\begin{tabular}{l | ccc | ccc}
      \hline
      \multirow{2}{*}{Method} & \multicolumn{3}{c|}{Tolerance 5cm} & \multicolumn{3}{c}{Tolerance 10cm}
      \\ \cline{2-4} \cline{5-7}
      & $F_1$ Score $(\%)$ & Accuracy $(\%)$ & Completeness $(\%)$ & $F_1$ Score $(\%)$ & Accuracy $(\%)$ & Completeness $(\%)\:\:$\\
      \hline
      MVSNet \cite{yao2018mvsnet}                  & 49.13        & \bf{83.40} & 35.27       & 56.22        & \bf{91.32} & 40.88\\
      MVSNet + Gipuma \cite{galliani2015massively} & 31.15        & 36.49      & 29.83       & 44.11        & 58.38      & 37.61\\
      R-MVSNet \cite{yao2019rmvsnet}               & 56.72        & 62.92      & 52.36       & 70.54        & 80.37      & 63.29\\
      DPSNet \cite{im2019dpsnet}                   & 30.74        & 28.77      & 35.63       & 44.61        & 44.78      & 46.66\\
      P-MVSNet \cite{Luo_2019_ICCV}                & 61.04        & 77.22      & 50.96       & 70.42        & 88.77      & 58.87\\
      A-TVSNet + Gipuma (Ours)                     & \bf{63.67}   & 55.66      & \bf{75.66}  & \bf{77.14}   & 72.66      & \bf{82.55}\\
      \hline
      \end{tabular}}
   \end{center}
   \caption{Comparisons of point cloud reconstruction results between different algorithms on ETH3D benchmark, larger is better.}
   \label{tab:eval_pc}
   \vspace{+0.2cm}
\end{table*}

\begin{figure*}
   \begin{center}
   \includegraphics[width=0.98\linewidth]{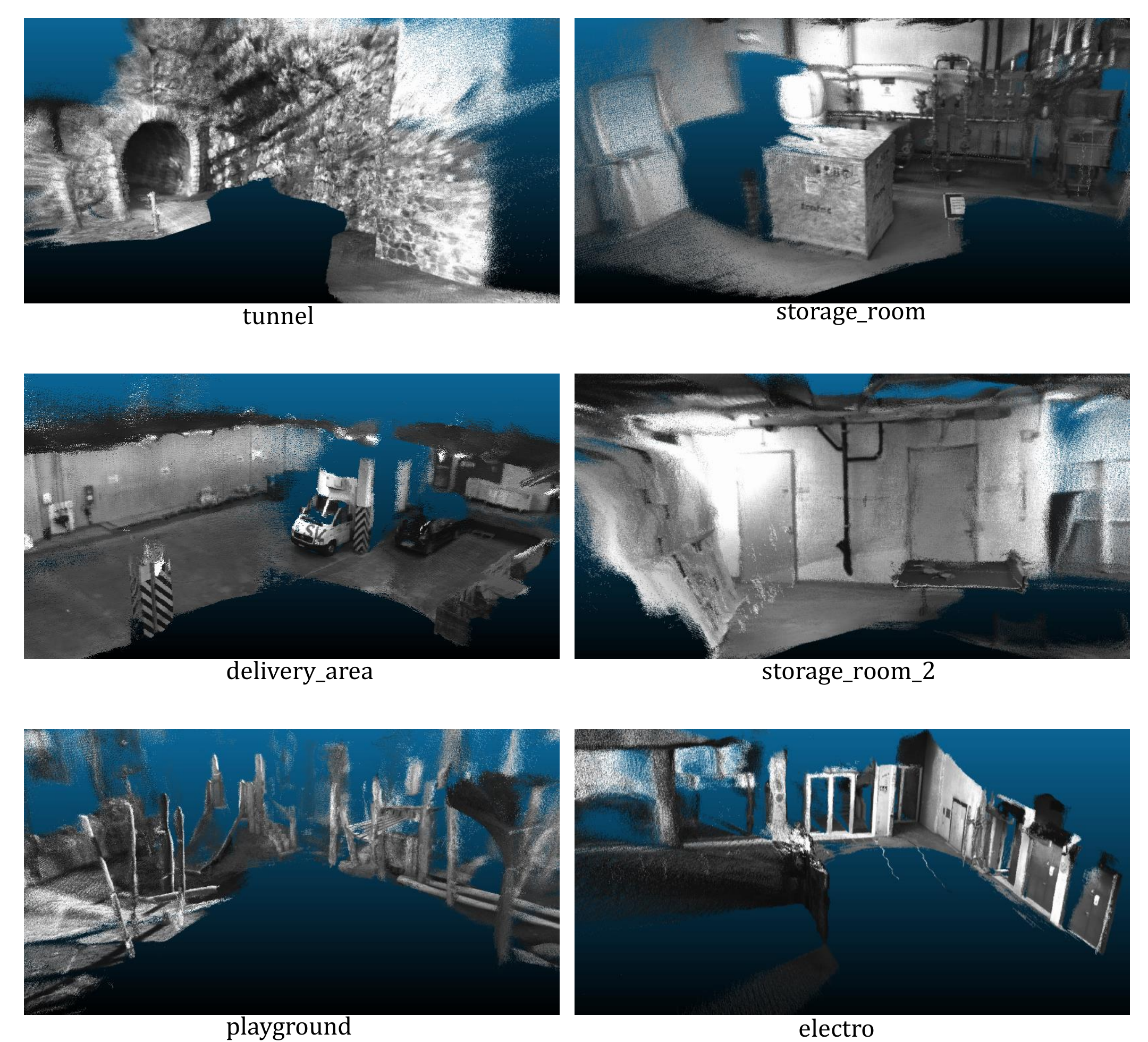}
   \end{center}
   \caption{Point cloud reconstruction results of ETH3D benchmark.}
   \label{fig:comparepc}
\end{figure*}

Since our A-TVSNet produces only depth map for each view, we use the same filtering and fusion strategy (but without normal consistency check) 
as Gipuma \cite{galliani2015massively} in order to generate point clouds from multiple depth maps. 
We compare our point cloud reconstruction results on the low-resolution many-view benchmark
of ETH3D dataset \cite{schoeps2017eth3d} as illustrates in Table~\ref{tab:eval_pc}. 
Our A-TVSNet shows better overall point cloud reconstruction quality (higher $F_1$ Score) than recent learning based methods.
Some point cloud reconstruction results are visually shown in Figure~\ref{fig:comparepc}.

\end{document}